\documentclass[runningheads]{llncs}
\usepackage{graphicx}
\begin{document}
\title{Automatic satellite building construction monitoring
}
%
%
\author{Insaf Ashrapov\inst{1}\orcidID{0000-0003-4938-0430} \and
Dmitriy	Malakhov\inst{3} \and
Anton Marchenkov\inst{2} \and
Anton Lulin \inst{3} \and
Dani El-Ayyass \inst{3}}

%
\institute{Moscow Institute of Physics
and Technology
(National Research University), Moscow, Russia
\email{intoff@mipt.ru}\\
\url{https://eng.mipt.ru/} \and
{Bauman Moscow State Technical University, Moscow, Russia \email{bauman@bmstu.ru} \url{https://bmstu.ru/mstu/English/}}
\and {Lomonosov Moscow State University, Moscow, Russia 
\email {info@rector.msu.ru}
\url{https://www.msu.ru/}}}

\maketitle 
\begin{abstract}
One of the promising applications of satellite images is building construction monitoring. It allows to control the construction progress around the world even in the locations that are hard to reach. One of the main hurdles of this approach is the interpretation of the image data. In this paper, we have employed several novel deep learning techniques to tackle the problem. Various image segmentation and object detection networks were combined into a unified pipeline, which was then used to determine the building construction progress.

\end{abstract}
\section{Introduction}
The initial problem is to determine the overall building construction progress. This task is crucial for the banking process purposes.

\textbf{Relevance of the problem}. There are several works and datasets with similar but rather different solutions. These include road and building detection \cite{albert2017using,Zang_2017,oehmcke2019detecting,prathap2018deep} and building heights estimation \cite{shi2020sar,Carvalho_2020,abdelrahim2017shadow}. Our solution was developed by using these results as a starting point.

 Current process relies on experts required to go construction sites in person. This happens due to the fact that provided videos and photos might have some unacceptable fraud artifacts (produced by GANs \cite{wang2019generative}, adversarial attacks \cite{akhtar2018threat,an2019unrestricted}, etc). In contrast, satellite images are free of such problems: experts do not need to visit the construction sites, and photos are free of fraud pre-processing. Besides, the price of satellite images continues to fall, making the usage of satellite images more affordable.

Recent developments in deep learning provide numerous opportunities in automatic satellite image analysis. The latest advances notably increased identification accuracy.  
Building construction consists of a variety of stages, each stage having its own unique properties. One thus needs to firstly determine which stage it is, and then to find out how much work is required to finish this stage.

\subsection{Building construction}
Building construction is the process of adding structures to areas of land, also known as real property sites. Typically, a project is instigated by or with the owner of the property (who may be an individual or an organization); occasionally, land may be compulsorily purchased from the owner for public use. The most important thing we should emphasize here is that the completion of each stage allows the building construction company to request a next portion of the loan from the bank. This is precisely why veracity and preciseness of building stage detection are highly important. 

As it was already mentioned, there exist different construction stages. All building stages and their order are shown in Fig.~\ref{fig0}:
\begin{figure}[hbt!]
\includegraphics[width=\textwidth]{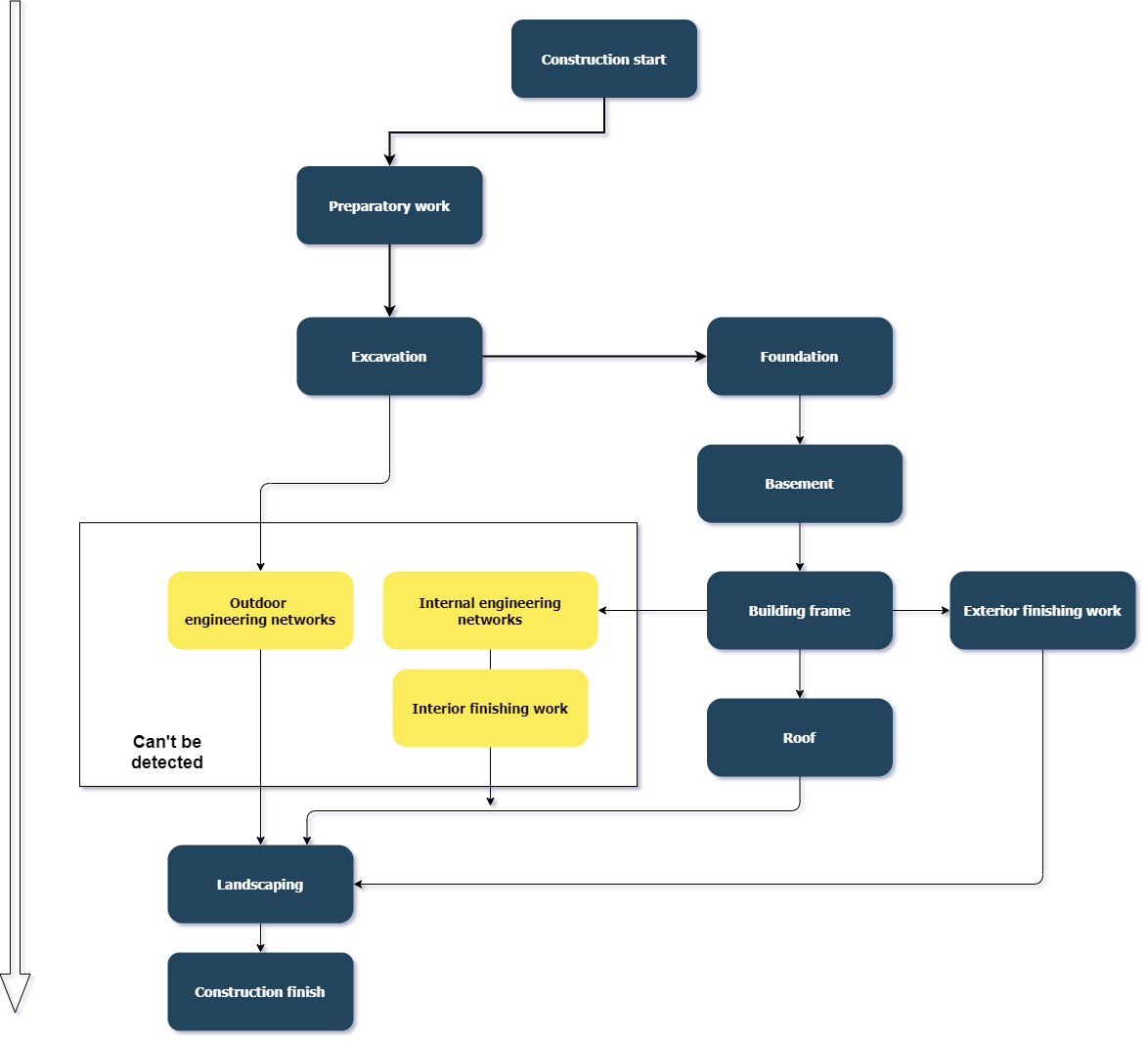}
\caption{Building stages order. Some of them might go in parallel.} \label{fig0}
\end{figure}

\section{Dataset}
To collect and obtain training datasets the initial large satellite images were cropped to smaller ones. Then after labeling the satellite photos, we validated each label one more time \cite{DBLP:journals/corr/abs-1811-03402}. Such an approach allowed us to obtain more stable results. 
The details of the labeling process are shown in Table~\ref{tab11}:

\begin{table}[hbt!]
\centering
\caption{Collected datasets of different building stages}\label{tab11}
\begin{tabular}{|l|l|l|l|l|}
\hline
 & \textbf{Stage} & \textbf{Label type} & \textbf{Total labels} & \textbf{Total Images} \\ \hline
0 & Preparatory work & Bbox & 2 180 & 1974 \\ \hline
1 & Excavation & Bbox & 2120 & 1974 \\ \hline
2 & Foundation & Bbox & 2 246 & 1974 \\ \hline
3 & Basement & Bbox & 2 088 & 1974 \\ \hline
4 & Building frame & Segmentation & 720 & 719 \\ \hline
5 & Roof/Completed house & Bbox & 132 195 & 2996 \\ \hline
6 & LandScaping & Segmentation & 240 & 203 \\ \hline
\end{tabular}
\end{table}

Taking into account that higher-quality satellite images cost dramatically more, we consider the 1 m/pixel image quality in the test. However, the images with the quality of 0.3-0.5 m/pixel were used for labeling. This happened due to the fact that one can always decrease image quality while training. Moreover, good image quality greatly facilitates the labeling process.

Table~\ref{tab_img} shows that decreasing the image quality by more than 2 times decreases overall metric only by 25\%. In other words we can train on \textit{good} quality images, but make inference on \textit{bad} quality images. We do so, because images with good quality cost much more.
\begin{table}[hbt!]
\caption{Model performance on different image quality}\label{tab_img}
\centering
\begin{tabular}{|l|l|}
\hline
\textbf{Image quality m/pixes} & \textbf{IOU, \%} \\ \hline
0.3 & 45 \\ \hline
0.8 & 36 \\ \hline
\end{tabular}
\end{table}

\section{Model Architecture}
\subsection{Segmentation pipeline}

In order to increase the training dataset size some augmentation techniques were employed \cite{Buslaev_2020}. The best results were achieved by using the horizontal flip method. Besides, some improvements were provided by brightness manipulations, horizontal shifts, and rotations. However, due to the nature of the satellite data, some valuable information is encoded in the orientation of images, thus making big rotations unacceptable in terms of model accuracy. As a final solution, we developed a single 5-fold model with horizontal flip TTA (test time augmentation) and symmetric Lovász-Softmax loss \cite{berman2017lovszsoftmax}. This pipeline was shown to perform well in the previous work \cite{DBLP:journals/corr/abs-1812-01429}. We used weight decay and Adam \cite{kingma2014adam}, as well. RAdam \cite{Liu2020On} turned out to be more stable than Adam, but often with worse results.

For the Landscaping building stage, we used segmentation as an inverse operation. Namely, we firstly segment by model non-landscaped area, and then by the proportion of non-landscaped surrounding area we determine the building stage progress.  

\textbf{Network}, we used vanilla Unet \cite{ronneberger2015unet} with some improvements. Inside each bottleneck, we used the Squeeze-and-Excitation network. The encoder was replaced by SE-ResNeXt-50 \cite{hu2017squeezeandexcitation}. It outperformed other encoders such as ResNets \cite{he2015deep} and even EfficientNet \cite{tan2019efficientnet}, which unfortunately consumed more GPU memory than expected and was unstable if input image size changed. The encoders were pre-trained on ImageNet \cite{russakovsky2014imagenet}. By doing so, one might noticeably improve the convergence speed and model quality \cite{iglovikov2018ternausnet,iglovikov2018ternausnetv2}.

\textbf{Pseudo Labeling}. Besides, acquiring additional unlabeled data with pseudo-labeling might improve the final results \cite{arazo2019pseudolabeling}. Hereby, we have applied it this way:

\begin{enumerate}
\item Typical training model pipeline described above
\item Applying a trained model on unlabeled images, which are neither a part of the train or test
\item Continuing to train the model on these new labels from pp.2, while validating on the same dataset as in pp.1
\item Repeat pp.2-3 up to 3 times
\item Results show increased metrics up to +3\% in terms of IOU (intersection over union)
\end{enumerate}

\subsection{Object detection pipeline}
So far highest model quality in tasks of object detection was provided by two-stage detectors. In contrast, one-stage detectors are faster and simpler but were lacking the accuracy of two-stage detectors this far. \cite{inproceedings}.

This is why we applied and tried regular Retina Net with ResNet50 with focal loss and NMS \cite{1699659} as post-processing to filter out unnecessary bounding boxes (bbox). Soft-NMS \cite{bodla2017softnms} gave worse results. In addition, to increase training size, we used the augmentation techniques described earlier in this work. The model structure is shown in Fig.~\ref{retina}:
\begin{figure}[hbt!]
\includegraphics[width=\textwidth]{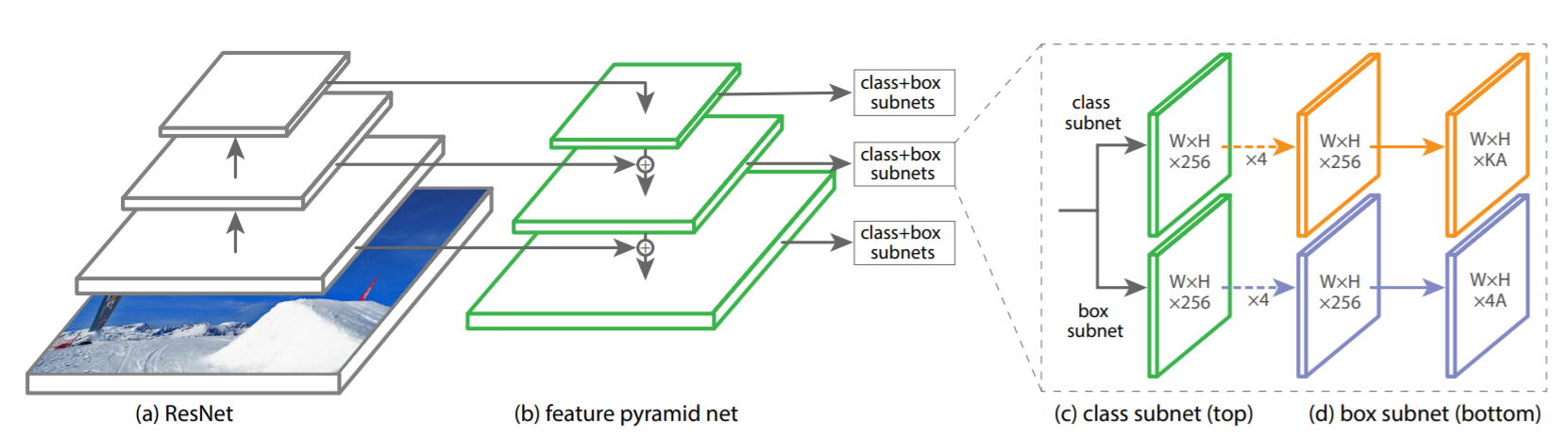}
\caption{The one-stage RetinaNet network architecture uses a Feature Pyramid Network \cite{inproceedings}} \label{retina}
\end{figure}

Faster R-CNN R101-C4 model \cite{detectron} was trained for comparison. However, it turned to be more than three times slower in training and inference without any noticeable improvements in final metrics.

\subsection{Building height estimation}
One of the crucial tasks to solve in building progress calculation is height estimation. Higher building means more progress done in building construction. It should be mentioned here that the building height we try to estimate does not depend on the topographic map.

Several methods were employed to solve such tasks:
\begin{itemize}
\item Multi-Task Learning of Height and Semantics from Aerial Images \cite{Carvalho_2020}. UNet with regression task using an RGB image as an input. For training DSM and DEM data are required, which are expensive and difficult to find. 
\item SAR Tomography at the Limit: Building Height Reconstruction Using Only 3 – 5 TanDEM-X Bistatic Interferograms. \cite{shi2020sar} Almost the same approach as pp.1 but uses several photos to detect a shift in images and heights map as a result.
\end{itemize}

\begin{figure}[hbt!]
\includegraphics[width=0.75\textwidth]{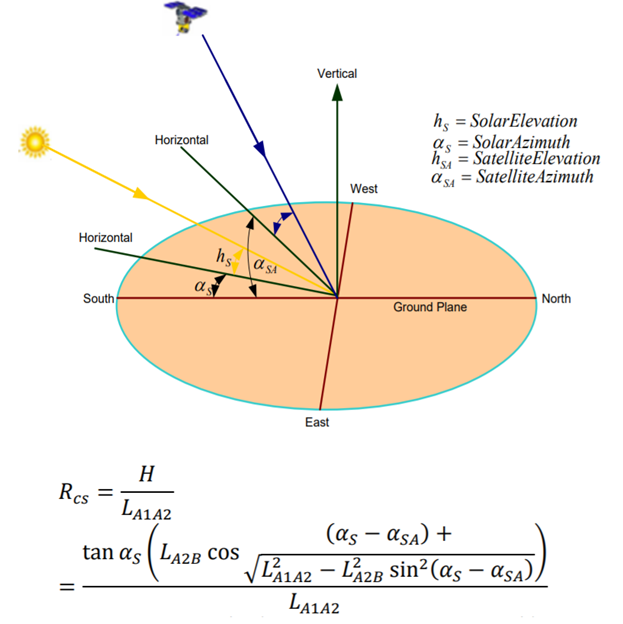}
\caption{Fundamental parameters of the sun and the satellite with the formula to determine building heights \cite{abdelrahim2017shadow}} \label{sun}
\end{figure}

However, both of them were unusable under given circumstances due to the lack of data. So we moved to shadow detection, using the length of the shadow to estimate the building height as stated by the formula in Fig.~\ref{sun}.

Firstly, to detect shadows we tried to use NIR (near infra-red range) image channel to detect shadow at some threshold. Unfortunately, the results were unacceptable, due to the trees shadows and water producing too much noise and False Positive results. 

Then, finally, we used the UNet-like architecture from the Segmentation pipeline to detect shadows and then applied the formula from Fig.~\ref{sun} to determine the desired building heights. We added that solution to the final pipeline.

\section{Results}
\subsection{Model quality and training set size}
\begin{figure}[hbt!]
\includegraphics[width=0.75\textwidth]{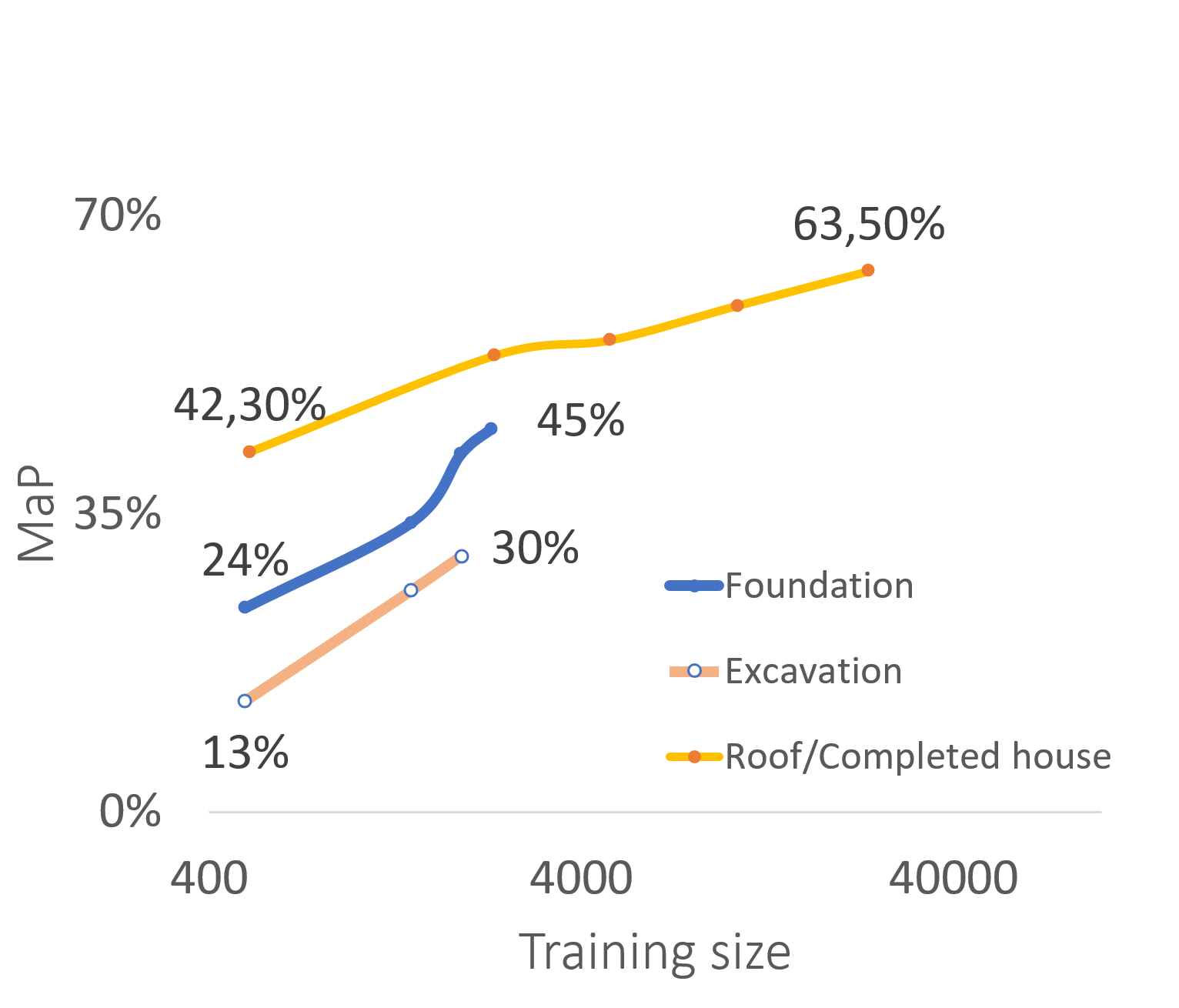}
\caption{Model performance depending on the training dataset size}
\label{gr1}
\end{figure}

For several building stages, we measure model metrics depending on training size. Even though we had enough objects only for one building stage, we can demonstrate a log-linear scale, as one can see in Fig. ~\ref{gr1}. Other building stages follow the same pattern, so we do not show them.

After 25 000 photos MAP (mean average precision) \cite{8594067} metrics stop to grow for the \textit{roof} stage, reaching a plateau. For other building stages, we believe that increasing the dataset size might help to improve the metrics. Similar log-linear dependency was shown in \cite{hestness2017deep}.

\subsection{Model Metrics}
\begin{figure}[hbt!]
\includegraphics[width=\textwidth]{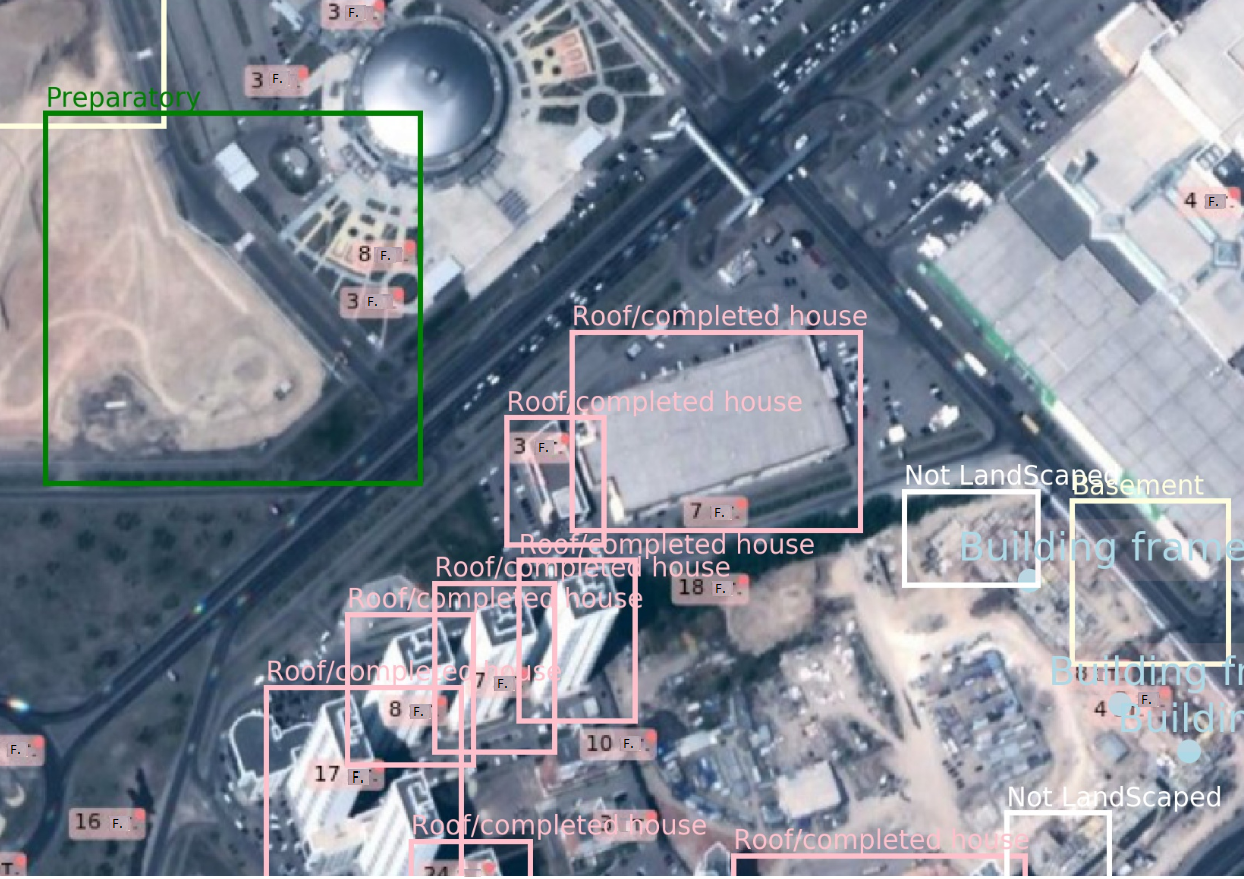}
\caption{Example of satellite building construction progress. Models inference.} \label{segm}
\end{figure}

The metrics scores achieved on the hold-out during the training process are shown in Table ~\ref{metrics}. Example of model inference on the satellite image shown in Fig.~\ref{segm}. The total completed building progress is estimated by classifying and detecting the building stages.
\begin{table}[hbt!]
\caption{Model metrics on different variety of tasks}\label{metrics}
\centering
\begin{tabular}{|l|l|l|l|l|}
\hline
 & Stage & Label type & Precision, \% & Recall, \% \\ \hline
0 & Preparatory work & Bbox & 38 & 35 \\ \hline
1 & Excavation & Bbox & 30 & 23 \\ \hline
2 & Foundation & Bbox & 45 & 50 \\ \hline
3 & Basement & Bbox & 25 & 27 \\ \hline
4 & Building frame & Segmentation & 85 & 77 \\ \hline
5 & Roof/completed house & Bbox & 74 & 70 \\ \hline
6 & LandScaping & Segmentation & 71 & 80 \\ \hline
7 & Building heights & Int & 65\% ($\pm$ 2 floors) & - \\ \hline
\end{tabular}
\end{table}

One should also notice that segmentation models were shown to perform quite well. For this reason, they might be considered as an alternative to object detection pipelines. Also, for segmentation, we need much less annotated images than for object detection.

\section{Conclusion}
In this work, we developed a first automated building construction monitoring solution. It provides sufficient quality to become a viable alternative to visiting construction sites in person. The technique described in this paper utilizes both segmentation and object detection models. Building shadows are used to identify the building height.  
Moreover, the models trained on high quality images can utilize much lower quality images on the inference stage without any noticeable quality decrease. Model performance depends on the training set size logarithmically.

\section{Acknowledgment}
 The authors would like to thank Open Data Science community \cite{ods} for many valuable discussions and educational help in the growing field of machine and deep learning. Also, big special thanks to PJSC Sberbank \cite{sber} for allowing to solve such tasks and for providing computational resources. 

%
%
%
%

\bibliographystyle{splncs04}
\bibliography{salt.bib}

\end{document}